\documentclass[conference]{IEEEtran}
\usepackage{array}
\usepackage{graphicx}
\usepackage{subfig}
\usepackage{algorithm}
\usepackage{algpseudocode}
\usepackage[english]{babel}
\usepackage{blindtext}
\usepackage{amsmath}
\usepackage{times}
\usepackage{epsfig}
\usepackage{graphicx}
\usepackage{amssymb}
\usepackage{multirow}
\usepackage{color}
\usepackage{cite}
\usepackage{booktabs}
\usepackage{paralist}
\usepackage{multirow}
\usepackage{url}

\newcommand{\Rmnum}[1]{\uppercase\expandafter{\romannumeral #1}}
\graphicspath{{pics/}}
\begin{document}

\title{Measuring Patient Similarities via a Deep Architecture with Medical
  Concept Embedding}

\author{\IEEEauthorblockN{\textbf{Zihao Zhu\IEEEauthorrefmark{1}, Changchang
      Yin\IEEEauthorrefmark{1}, Buyue Qian\IEEEauthorrefmark{1}, Yu Cheng\IEEEauthorrefmark{2}, Jishang
      Wei\IEEEauthorrefmark{3}, Fei Wang\IEEEauthorrefmark{4}}}
  \IEEEauthorblockA{
    \IEEEauthorrefmark{1}Xi'an Jiaotong University,
    \IEEEauthorrefmark{2}Microsoft AI \& Research\\
    \IEEEauthorrefmark{3} HP Labs, \IEEEauthorrefmark{4} Weill Cornell Medical School\\
    }}

\maketitle

\begin{abstract}
Evaluating the clinical similarities between pairwise patients is a fundamental problem in healthcare informatics. A proper patient similarity measure enables various downstream applications, such as cohort study and treatment comparative effectiveness research. One major carrier for conducting patient similarity research is the Electronic Health Records(EHRs), which are usually heterogeneous, longitudinal, and sparse. Though existing studies on learning patient similarity from EHRs have shown being useful in solving real clinical problems, their applicability is limited due to the lack of medical interpretations. Moreover, most previous methods assume a vector based representation for patients, which typically requires aggregation of medical events over a certain time period. As a consequence, the temporal information will be lost. In this paper, we propose a patient similarity evaluation framework based on temporal matching of longitudinal patient EHRs. Two efficient methods are presented, unsupervised and supervised, both of which preserve the temporal properties in EHRs. The supervised scheme takes a convolutional neural network architecture, and learns an optimal representation of patient clinical records with medical concept embedding. The empirical results on real-world clinical data demonstrate substantial improvement over the baselines. We make our code and sample data available for further study.
\footnote{\url{https://github.com/yinchangchang/patient_similarity}}
\end{abstract}


\section{Introduction} \label{sec:introduction}

Patient similarity learning has been identified as one of the key techniques for healthcare transformation. During the past decade, Electronic Health events (EHR), including diagnosis codes, lab results, prescription data, are becoming readily available for a huge amount of patients. This makes EHR a valuable resource for evaluating the clinical similarities between pairwise patients. Patient similarity, which measures how similar a pair of patients are according to their historical information under a specific clinical context, will be the enabling technique for making various healthcare applications possible, such as cohort analysis, case based reasoning, treatment comparison, disease sub-typing, and personalized medicine. In addition, learning patient similarity is a fundamental problem in evidence based medicine, which has been identified as one of the major thrust areas for transforming healthcare and improving the quality of delivery of care.

\textbf{Motivation.} One of the key challenges to derive patient similarity measure is how to represent the medical events of patients effectively without loss of information. Since a great deal of healthcare analytics applications critically rely upon patient similarity, the similarity measures need to be both clinically effective and accurate. Though important, there are only a handful studies on patient similarity learning \cite{Cheng2016Risk}\cite{sun2012supervised}. Existing methods have successfully derived the similarity measure from EHR data through mapping the medical events into vector spaces, however, their applicability is limited due to the lack of convincing explanations for patient representations in medical domain. There has been some existing work on applying patient similarity to various applications in medical literatures. However, there are still significant challenges on learning effective patient similarities, which, to our knowledge, have not been systematically addressed. \emph{(i) Temporal-Sensitivity:} Temporal information is important to medical events, and is crucial to understand the dynamics of medical expressions. \emph{(ii) High Dimensionality and Sparsity:} EHR includes a wide range of data (such as diagnosis, medication, lab test) and a large number of possible medical events (over ten thousands of diseases and medications), so that EHR data is usually represented in a high dimensional space. Besides, EHR data is also very sparse, since a record exists if and only if the patient pays a visit to a specific clinical institute, for a particular condition. \emph{(iii) Limited interpretability:} Due to the complexity of medical data, existing patient representation models are often weak at the perspective of clinical interpretations, which if addressed would significantly widen their applicability.

\textbf{Proposal.}  Taking into account all challenges mentioned above, inspired by the idea of words embedding \cite{mikolov2013distributed}, we propose a method to represent patients and derive a similarity measure based on it. Unlike previous methods that model each medical event as a binary event vector over time (one if the medical event happened and zero otherwise), we derive a fixed-length vector representation from EHRs by medical concept embedding. In text mining, a particular word can be predicted based on the context around it \cite{mikolov2013efficient}\cite{mikolov2013distributed}. Similarly, events happened before and after a specific medical event can be viewed as its medical context, which may be used to make event predictions in medical domain. Based on the medical context, each event is compressed into a given length vector with medical concept embedding. Similar to the word embedding \cite{mikolov2013distributed}, the event embedding presented in our model hold its natural medical concept. Furthermore, we adjust the range of context, with respect to the specific conditions of a medical event, to achieve an event embedding with temporal information. By stacking all event embedding vectors together, each patient is then represented as an embedding matrix. Note that, compared to describing patients using binary event vectors, the embedding extracts clinical features of a patient from EHRs and represent them in a reasonable dimension, resulting a natural dense embedding matrix for every patient.

Based on the embedding matrix representation of patients, we propose two methods, supervised and unsupervised, to derive the similarity measures. Note that the number of medical events varies from patients to patients, and both the supervised and unsupervised approaches are required to measure the similarity between matrices with different dimensions. As for the unsupervised method, we adopt the $RV$ coefficient \cite{josse2013measures} and $dCov$ coefficient \cite{szekely2007measuring}, respectively, to measure linear and non-linear relations between pairwise patients based on the embedding matrix. In the supervised model, we measure the patients similarity using the Convolutional Neural Network (CNN), where the deep medical embedding is obtained from the intermediate convolutional feature maps. With the given number of convolutional filters, an event embedding matrix is mapped to a fixed-length feature vector. The deep medical concept embedding contributes to improved patients similarity measures. We shall later in the paper make a comparison amongst different types of patient representations, including the binary event matrix representation.

\textbf{Empirical Study.} Patient cohort study is the most effective way to analyze the causes, treatments, and outcomes of diseases. To evaluate the representations we proposed, we conduct a cohort analysis based on the obtained measures of patient similarity. Our model is tested on a real-world EHR dataset containing a wide range of medical events over a long time period. The experimental results demonstrate the effectiveness of our model in measuring patient similarity.

\textbf{Contributions.} Our work makes the following distinctive technical contributions:
\begin{itemize}
\item We adopt a state-of-the-art distributional representation model to project medical events to fixed-length vectors, which are then used to measure patient similarity.
\item We effectively extract the low-dimensional and dense representation for patients from EHR data, with the temporal information preserved.
\item We propose two solutions for patient similarity Learning, unsupervised and supervised. This makes our framework applicable to most similarity-related applications in healthcare analytics.
\end{itemize}

The rest of this paper is organized as follows. Section \ref{sec:related work} introduces related studies. Details about our model are presented in Section \ref{sec:method}. The experimental results are reported in Section \ref{sec:experiments}, and Section \ref{sec:conclusions} concludes.
\section{Related Work} \label{sec:related work}
In this section, we first review some related work on evaluating the clinical patient similarities, and then review some relevant problems associated with deep learning.

\subsection{Patient Similarity}
In healthcare informatics domain, there are a lot of works focusing on patient similarity. For example, \cite{chan2010machine}
proposed a patient similarity algorithm named SimSvm that uses Support Vector
Machine(SVM) to weight the similarity measures. \cite{wang2012medical} proposed
a patient similarity based disease prognosis strategy named SimProX. This model
uses a Local Spline Regression (LSR) based method to embed these patient events into an intrinsic space, then measure the patient similarity by the Euclidean distance in the embedded space. These methods do not take the temporal information into consideration when evaluating patient similarities. Wang \cite{wang2012towards} presented an One-Sided
Convolutional Matrix Factorization for detection of temporal patterns. Cheng \cite{Cheng2016Risk,chenips} proposed an adjustable temporal fusion scheme using CNN-extracted features. Based on patients similarity, plenty of applications are enabled. In \cite{ng2015personalized}, Ng provided personalized predictive healthcare model by matching clinical similar patients with a locally supervised metric learning measure. \cite{kasabov2010integrated} proposed Integrated Method for Personalised Modelling (IMPM) to provide personalised treatment and personalised drug design.

There are many research have been conducted on clustering patients based on
machine learning. In order to rate patients health perceptions, Sewitch
\cite{sewitch2004clustering} make cluster analysis using k-means to identify the
patients groups based on the discovering the multivariate pattern. To capture
underlying structure in the history of present illness section from patients
EHR, Henao \cite{henao2013patient} proposed a statistical model that groups
patients based on text data in the initial history of present illness (HPI) and
final diagnosis (DX) of a patient’s EHR. For human disease gene expression,
Huang \cite{huang2013spectral} presented a new recursive K-means spectral
clustering method (ReKS) to efficient cluster human diseases. Most of these
research have demonstrate effectiveness of their model with real-world
experiments, that convinces us of the applicability of clustering patients on
cohorts discovering.

\subsection{Embedding Learning and Semantic Matching}
One of the most important components in our patients similarity
measure is deep distributional medical concept embedding.
Distributional representations has gone through the long evolution,
and shows state-of-the-art results in many fields recently.
\cite{mikolov2013efficient,mikolov2013distributed} proposed
continuous Bag-of-Words model and Skip-gram model to represent words
in vector space. The word representations using neural networks
provide state-of-the-art performance on measuring syntactic and
semantic word similarities. Many works as well as ours are inspired
by the words embedding with neural networks.
\cite{kiela2014learning} learned image embedding by concatenating
skip-gram linguistic representation vectors with visual concept
representation vectors. \cite{severyn2015learning} encoded a
query-document pairs into discriminate feature vectors using
distributional sentence model. Similar embedding also has been used
in other applications \cite{10.1093/jamia/ocx090,ehr-gan}. Our model
achieves the goal of embedding patients clinical features in the
dense matrices with modest dimensionality. This

With medical concept embedding, we look forward to calculating the similarity
amongst patients according to their EHRs. Considering the representations of
patient medical events do not have a common time dimension, we cannot compare
the patient event matrix directly. \cite{kherif2003group} provided a relevant
similarity measures between temporal series of brain functional images belonging
to different subjects. Similar to \cite{kherif2003group}, we adopt the RV
coefficient to measure patient similarities. Note, however, that this
coefficient only considers linear relationships between two data sets. To do
more systematic research on measuring similarity of patient, our model also
measures non-linear correlation between two patients using dCov
coefficient. Apart from those unsupervised approach, we adopt the supervised learning method. We modify the Convolutional Neural Network(CNN) to derive the similarity scores for pairs of patients. The Convolutional networks models which are originally invented for image processing have wide applications in other
domains. \cite{severyn2015learning,hu2014convolutional} and
\cite{lu2013deep} respectively obtains the continuous representations of the
sentences or short texts by a convolutional deep network, then the similarity
can be effectively established.

\section{The Proposed Method} \label{sec:method}
Accessing patients similarities in EHR data is a very challengeable
task. In this section, we will fist propose to learn the ontextual
embedding of each medical concept. Then we provide an unsupervised
method to estimate the similarity score, which takes the learned
medical concept as the input. After that we exploit an architexture
building with convolutional neural network to measure the similarity
of pair patient records with some supervision encoded.

\subsection{Contextual Embedding of Medical Concepts}
Our goal in this step is to get the contextual embedding of each medical concept from patient EHR, which provide a better representation for medical concepts than general one-hot encoding representation. By ``context” around a medical concept A we really mean the medical events happening before and after A within the patient EHR corpus. For each patient, by concatenating all medical events in his/her EHR according to their happening timestamps (for events with the same timestamp we do not care about the order), we obtained a “paragraph” describing the historical condition of him/her. So the context around a specific medical event is similar to the context around a word in a paragraph. How to derive effective word representations by incorporating contextual information is a fundamental problem in Natural Language Processing (NLP) and has been extensively studied. One recent advance is the “Word2Vec” technique \cite{mikolov2013efficient} that trains a two-layer neural network from a text corpus to map each word into a vector space encoding the word contextual correlations. The similarities (usually cosine distance) evaluated in such embedded vector space reflect the contextual associations (e.g., words A and B with high similarity suggests they tend to appear in the same context).

In NLP, the context around each word is usually identified as the adjacent words before and after it. In Word2Vec such context is defined by a sliding window around each word and the length of the window reflects the scope of the context. In EHR, as there is a timestamp associated with each medical concept, we do not just want to consider the relative positions when defining the context, but consider the actual timestamps. For example, we may want to treat event B happened one year after A differently comparing to event B happened one week after A. Another factor we need to consider is the context scope around each event, i.e., the length of the sliding windows. In Word2Vec models for NLP every word is assigned with the same window length. In contrast, for EHR, we may want medical concepts related to chronic conditions to have larger scopes while acute condition concepts to have smaller scopes. Moreover, because of the variabilities among individual patients, the scope for the same event could be different for different patients. Therefore we propose an adaptive way to determine the window length for an event in the EHR of a specific patient. Our heuristic is that chronic conditions are more likely to appear repeatedly in a patient’s EHR and thus have higher frequency, and acute conditions will be less frequent. Then for medical event $i$ and patient $p$,
\begin{equation}
L(i,p) = f(i,p)*a + \theta
\label{eq:w2v}
\end{equation}
where $f(i,p)$ is the frequency of event $i$ in the EHR of patient $p$. $a$ and $\theta$ are constants.

\subsection{Temporal Patient Representation}
After the medical concepts embedding step, We expect that the medical concept representations learned by Skip-gram will show
similar properties so that the concept vectors will support clinically meaningful vector additions. A straitfoward representation of a patient $p$ will be as simple as converting all medical concepts in his medical history to medical concept vectors, then summing all
those vectors to obtain a single representation vector. However, this representation will loss the temporal information. Instead, we utilize a temporal representation:
the records of each patient $p$ is represented as a matrix $\mathbf{X}$ with dimension $d \times N_{p}$, where $d$ is the fix embedding dimension and $N_{p}$ is the total number of visit patient $p$ has. A single representation vector of one visit is obtained by umming all the medical vectors in that visit. Usually, $N_{p}$ varies from patient to patient.
Given two patients $p_{a}$ and $p_{b}$, calculating the similarity between the record $\mathbf{X}_{a}$ and $\mathbf{X}_{b}$ is not that intuitive. We propose the method in the following sections.

\begin{figure*}
    \centering
    \includegraphics[width = 0.7\textwidth]{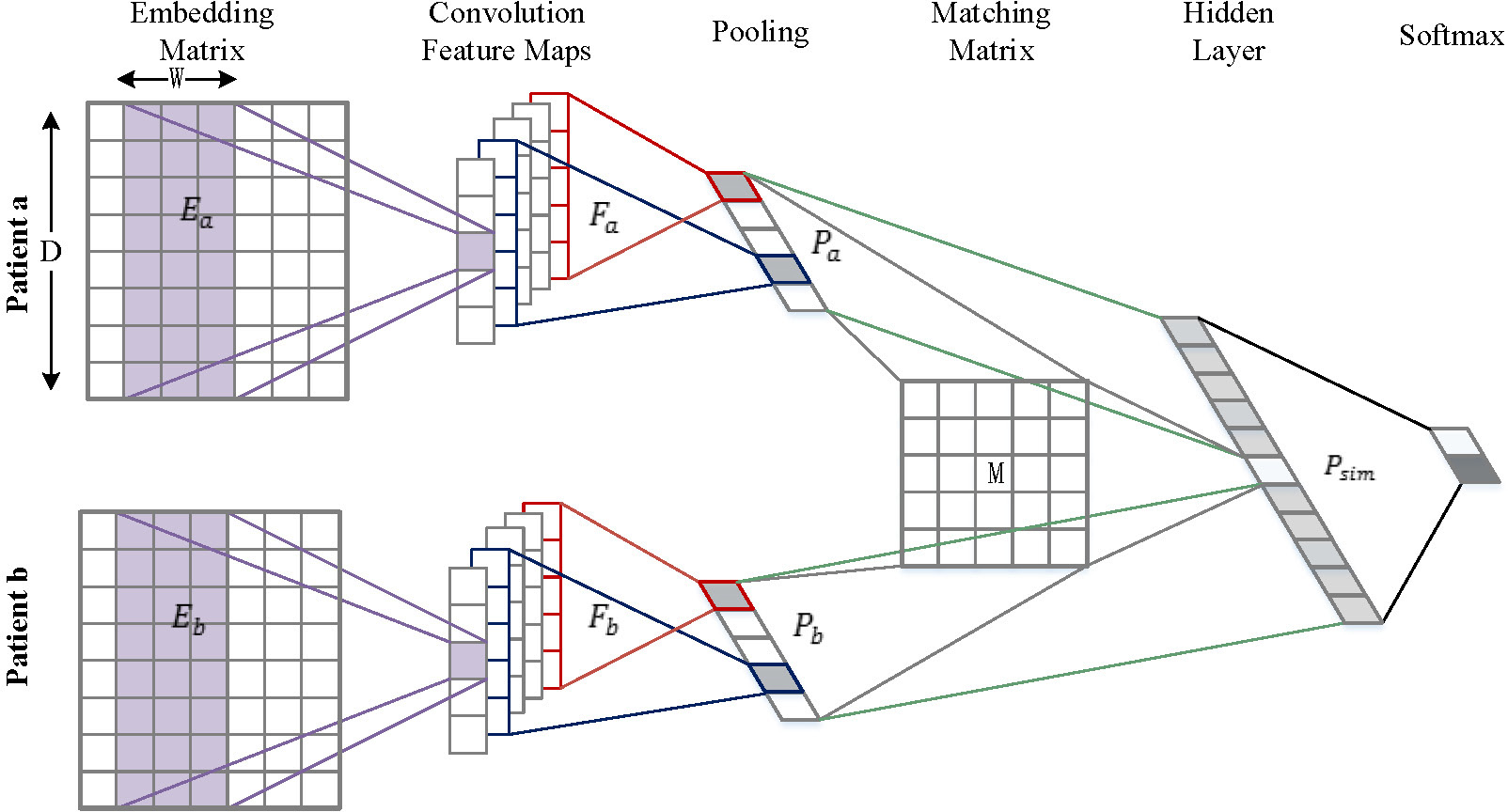}
    \caption{The overall framework of supervised patient similarity matching. To train the sigular neural network, embedding matrices of pairs of patients $E_a$,$E_b$ passed through convolutional filters are mapped into feature maps. We build the deep embedding patients representations $P_a$,$P_b$ for patients by pooling patients feature maps into the intermediate vectors. With the rich feature vectors we learn a symmetrical similarity matirx $M$ for measuring the distance between patient $a$ and $b$.}
    \label{fig:petscan}
  \end{figure*}

\subsection{Unsupervised Patient Similarity}
In order to calculate the similarity score based on the patient temporal representation, we provide two alternatives. The first one is to utilize RV coefficient \cite{} and dCov efficient{} to estimate the similarity over the pair of temporal patient representation. In particularly, given two matrix representations $\mathbf{X} \in \mathcal{R}^{n \times k}$ and $\mathbf{Y} \in \mathcal{R}^{m \times k}$, the RV coefficient is defined as:
\begin{equation}
\rm{RV}(\mathbf{X},\mathbf{Y}) = \frac{tr(\mathbf{X}\mathbf{X}^{`}\mathbf{Y}\mathbf{Y}^{`})}{\sqrt{tr(\mathbf{X}\mathbf{X}^{`})^{2}tr(\mathbf{Y}\mathbf{Y}^{`})^{2}}}
\label{eq:rv}
\end{equation}

For the dCov efficient, let's first define the empirical distance covariance:
\begin{equation}
\rm{dCov^{2}_{n}}(\mathbf{X},\mathbf{Y}) = \frac{1}{n^{2}} \sum_{i,j=1}^{n} (d_{ij}^{X}-d_{i.}^{X}-d_{.j}^{X}+d_{..}^{X})(d_{ij}^{Y}-d_{i.}^{Y}-d_{.j}^{Y}+d_{..}^{Y})
\label{eq:dcov}
\end{equation}
where $d_{ij}()$ is the Euclidean distance between sample $i$ and $j$ of random vector $\mathbf{x}_{i}$, $d_{i.}=\frac{1}{n}\sum_{j=1}^{n}d_{ij}$, $d_{.j}=\frac{1}{m}\sum_{i=1}^{n}d_{ij}$, $d_{..}=\frac{1}{n}\sum_{i,j=1}^{m,n}d_{ij}$. The empirical distance correlation (dCov efficient) is defined:
\begin{equation}
\rm{dCor^{2}_{n}}(\mathbf{X},\mathbf{Y}) = \frac{dCov^{2}_{n}(\mathbf{X},\mathbf{Y})}{\sqrt{dCov^{2}_{n}(\mathbf{X},\mathbf{X})dCov^{2}_{n}(\mathbf{Y},\mathbf{Y})}}
\label{eq:dcoe}
\end{equation}

\subsection{Measure Similarities with Supervision}
In order to add some supervision to this procedure, we proposed a deep learning model. The idea is derived from semantic matching problem in NLP, which aims to determine a matching score for two given texts. Deep learning approach has been applied to this area and most of the models conducts the matching through creating a hierarchical
matching structure built on convoluational neural nets (ConvNets).
The architecture of our model for measure patient pairs
is presented in Figure \ref{fig:petscan}.
The models based on ConvNets learn to map input patient representation to vectors, which
can then be used to compute their similarity. These are then used
to compute a patient similarity score, which together with
the representation vectors are joined in a single representation.

In the following we describe how the intermediate representations
produced by the ConvNets model can be used to compute patient
similarity scores and give a brief explanation of the remaining
layers, e.g. hidden and softmax, used in our network.

\textbf{Single Convolution feature maps}: The aim of the convolutional layer is to extract effective patterns, i.e., discriminative medical concept sequences found within the input record that are common throughout the training instances. In general, let $\mathbf{x}_i \in \mathbb{R}^d$ be
the $d$-dimensional event vector corresponding to the $i$-th time items. A one-side
convolution operation involves a filter $w \in \mathbb{R}^{d \times
h}$, which is applied to a window of $h$ event features to produce a
new feature. For example, a feature $c_i$ is generated from a window
of events $x_{i:i+h-1}$ is defined by:
\begin{equation}
c_{i} = f(\mathbf{w} \centerdot
x_{i:i+h-1} +b)
\label{eq:convnet}
\end{equation}
where $b \in \mathbb{R}$ is a bias term and $f$ is
a non-linear function (we use rectification (ReLU)).

\textbf{Pooling}:
The output from the convolutional layer (passed through the activation
function) are then passed to the pooling layer, whose goal is to aggregate the information and reduce the representation. This filter is applied to each possible window of features in the
event matrix $\{x_{1:h}, x_{2:h+1},...,x_{n-h+1:n}\}$ to produce a
feature map $ \mathbf{c} = [c_{1},c_{2},...,c_{n-h+1}]$, where
$\mathbf{c} \in \mathbb{R}^{n-h+1}$. We then apply a max pooling
over the feature map and take the average value $\hat{c} =
max\{\mathbf{c}\}$. The idea is to capture the most important
feature one with the highest value for each feature map.

\textbf{Matching Matrix}:
Given the output of our basic for processing patient records, their resulting vector representations $\mathbf{x}_a$ and $\mathbf{x}_b$,
can be used to compute a record-record similarity score. We follow
the approach of \cite{bordes2014open} that defines the similarity between $\mathbf{x}_a$ and
$\mathbf{x}_b$ vectors as follows:
\begin{equation}
\rm{sim}(\mathbf{x}_a,\mathbf{x}_b) = \mathbf{x}_{a}^{T} \mathbf{M} \mathbf{x}_{b}^{T}
\label{eq:matching}
\end{equation}
where $\mathbf{M} \in \mathbb{R}^{m \times m}$ is a similarity matrix. The similarity matrix $\mathbf{M}$ is a symmetrical parameter of
the network and is optimized during the training.

\textbf{Softmax}: The output of the penultimate convolutional and pooling layers
is flattened to a dense vector $\mathbf{x}$, which is passed to a fully connected softmax layer. It computes the probability distribution over the labels.

\subsection{Optimization}
For different tasks, we need to utilize different loss functions
to train our model. Taking regression as an example, we can
use square loss for optimization:
\begin{equation}
\mathcal{L}(S_1,S_2,y) = (y-M(S_1,S_2))^{2}
\label{eq:loss}
\end{equation}
where $y \in R$ is the real-valued ground-truth label to indicate
the matching degree between $S_1$ and $S_2$.

All parameters of the model, including the parameters of
word embedding, neural tensor network, spatial RNN are
jointly trained by back-propagation and Stochastic Gradient
Descent. Specifically, we use AdaGrad \cite{duchi2011adaptive}
on all parameters in the training process.

\textbf{Regularization}
For regularization we employ dropout on the
penultimate layer. Dropout
prevents co-adaptation of hidden units by randomly
dropping out—i.e., setting to zero—a proportion
$p$ of the hidden units during foward-back-propagation.

\section{Experiments and evaluation}
\label{sec:experiments}
In this section, we evaluate our framework on a real clinical EHR dataset. We carry out the cohort studies by selecting several chronic diseases
associated with a range of comorbidities. There are some reasons for our cohort
selection. First, they are frequently occurred diseases being extensively analyzed in healthcare applications. Second, these diseases are highly associated with each other, and their combination presents many diagnostic challenges. More importantly, due to the long period progression path of those disease, there are a great deal of temporal information embedded in the medical events. Many of medical research
based on machine learning ignored the temporality while our model effectively
extract those features and enrich the patients representations. Based on
patients clinical similarities derived from these representations, we group
patients into clusters by some classical clustering algorithms. As we focus on
matching similar patients, the clustering evaluations verify the effectiveness
of our model.

As testing our model on the real world EHRs, we demonstrate that our method can
effectively represent patients without sacrificing temporal information. With the distributional continuous representations, we apply deep neural networks to derive
measure of similarities amongst patients in the datasets. We then make use of the
similarity matrix to group patients. For the evaluations shown in the results, we are convinced that the deep medical event embedding achieves a significant
improvement in patients representations.

Further more, we demonstrate the robustness of our model in the cohort
studies. As mentioned in \cite{song2010observational}, the primary disadvantage
of medical cohort study is the limited control the investigator has over data
collection. The existing data may be incomplete, inaccurate, or inconsistently
measured between subjects \cite{browner1988designing}. As a result, we process
patients EHR for constructing two kinds of data sets. One covers the whole complete
patient events for global features analyzing. On another data set, we remove
particular events labeled as cohort identifers from patients EHR to provide
more natural setting in clinical cases. We systematically analyze the performance of our model in
the above two settings, and draw some conclusions through our result discussions.

\subsection{Datasets}
Our model is trained on a real world longitudinal EHR database of 218,680 patients for the course of over four years. According to the reasons presented at the beginning of this section, we select four patient cohorts from the EHR data, namely, Chronic Obstructive Pulmonary Disease (COPD), Diabetes, Heart Failure, and Obesity.

Table \ref{tab:Cohort-Table} provides a summary of the patient cohorts used in our experiment. Each cohort consists of a set of case patients who are confirmed with one of the four
diseases according to their medical diagnosis, and each patient comes with a set of
medical events including diagnosis and medications. In each patient encounter,
we use the International Classification of Disease-Version 9 (ICD-9) codes to
denote the diagnosis of diseases that a patient suffers from. All the clinical events about
medications are pre-processed to normalize the descriptions based on brand
names and clinical dosages.

\begin{table}[!htbp]
  \centering
  \begin{tabular}{lcc}
    \toprule
    Cohorts& \# Patients& \# Events\\
    \midrule
    COPD& 2,000& 247,043\\
    Diabetes& 2,000& 259,074\\
    Obesity& 2,000& 211,496\\
    Heart Failure& 1,135& 165,254\\
    \midrule
    Total& 7,135& 882,867\\
    \bottomrule
  \end{tabular}
  \caption[Summary-EHRs]{Summary of EHR datasets for patients clustering.}
  \label{tab:Cohort-Table}
\end{table}

We construct datasets with medical events collected from patients who were confirmed of
having the disease by medical experts. We develop the criteria that any patients
presented in the datasets has at least forty events. The requirement is set to
ensure that each test case has minimum events of clinical history that could be used
in reasonable analytics tasks in healthcare. Also, to enable distinctly cluster without overlapping
among cohorts, we remove patients who suffers from more than one disease in the
cohort list. Finally, there are 8,000 remaining patients and 6,064
distinct clinical events. Medical event appearing in more than
90\% of patients or present in fewer than five patients are removed from the
datasets to avoid biases and noise in the learning process.

In the following experiments, we create two datasets: DATASET-\Rmnum{1} uses the
complete patients events while DATASET-\Rmnum{2} reserves historical events except
those labeled as cohort identifiers. On DATASET-\Rmnum{1}, we split the dataset into
training and test sets with same number of patients, and other patients left for
validation. As for DATASET-\Rmnum{2}, we construct the data sets in accordance with
DATASET-\Rmnum{1}. A few of patients are filtered out because of the limited number of
their medical events. Table \ref{tab:two-datasets} summaries the two datasets.

\begin{table}[!htbp]
  \centering
  \begin{tabular}{lcc}
    \toprule
    Data& \# Patients& \# Events \\
    \midrule
    DATASET-\Rmnum{1}&& \\
    \quad TRAIN& 3,211& 396,072 \\
    \quad TEST& 3,210& 399,804 \\
    \quad DEV& 714& 86,991 \\
    \midrule
    DATASET-\Rmnum{2}&& \\
    \quad TRAIN& 3,083& 373,145 \\
    \quad TEST& 3,080& 377,287 \\
    \quad DEV& 685& 81,392 \\
    \bottomrule
  \end{tabular}
  \caption{Summary of modeling datasets.}
  \label{tab:two-datasets}
\end{table}

\subsection{Medical concept embeddings}
\label{sec:experiments-2}
We use word embeddings to represent each medical event as a vector.
We run \texttt{word2vec} on the datasets containing 218,680 patients with around
16.9 million medical event records. To learn the embeddings, we choose the Bag of Words
model with window size setting to 20 and events filtering with frequency less
than 5. The dimensionality of our embedding vectors $d$ is set to 20, 30, 50, 200,
500, respectively, for the comparison purpose, and after a serial practices we select 50 as
medical event dimension according to the best performance. Finally, the resulting
event matrix covers around 8,000 events which are presented using 50-dimensional
vectors, and the event matrix contains all of medical features of patient. Next, we shall
discuss how to use them for representing individuals and measuring their
distances.

\subsection{Experimental Settings}
\label{sec:experiments-3}
The parameters of our deep learning were as follow: the width of the convolution
filters $w$ is set to 5, 10, 15, 20, 25, and the number of convolutional feature
maps $m$ takes on 50, 100, 150, 200. We use stochastic gradient descent to
optimize the model's parameters. We train the model with 50 examples of shuffled
mini-batches. We adopt non-linear rectification (ReLU) activation function and a
simple max-pooling to achieve the intermediate representations. With regards to
overfitting issue we add dropout regularization with dropout rate setting to
0.5.

To optimize our deep features embedding, we conduct experiments using several
different parameters sets $\theta = \{d,w,m\}$, which vary in size of
\texttt{word2vec} embedding dimension, convolution filters width, and the number
of convolutional feature maps. In oder to find optimal set of parameters, we
compare the performance of clustering with only one variable of $d$,$w$,$m$
varies.

We implement the clustering base on following representations:
\begin{inparaenum}[(1)]
\item One-hot representation. Patient is represented as an event matrix. The
  matrices are composed of medical event columns, the dimension of which is set
  to 8,000, or the number of distinct medical events. The event matrix is
  naturally sparse, but it simplifies patients descriptions.
\item ``Shallow'' embeddings. As described in section \ref{sec:experiments-2},
  we make progress in patients representations with medical event embedding by
  \texttt{word2vec}. Similar to One-hot representation, we represent patients as
  matrices, but denser and lower dimensional.  The dimension of matrix columns
  has been reduced, with setting from 50 to 800.
\item Deep embeddings. To achieve a deep representation, we combine CNN with
  distributional medical events embeddings from \texttt{word2vec}. Based on
  above event matrix representations, patients features are filtered through the
  convolutional layer of neural network. Feature maps that represent patients
  clinical characteristics are then used to measure patients distances.
\end{inparaenum}

With generated representations of each patient, we firstly calculate the
similarity amongst all the test patients. Then, we group patients cohorts by
matching pairs of patients according to their similarity. $\mathtt{Kmeans}$ and
$\mathtt{Active \ PCKMeans}$ \cite{basu2004active} are adopted for grouping
patients based on the first two representations. Also, we compare our model with
another metric learning algorithm that have shown state-of-the-art results in
clustering. Besides deep neural networks we have applied to learn patients
features, we present other two unsupervised methods for calculating patients
distances as complementary. Specifically, we use $RV$ and $dCov$ coefficient to
calculate correlations of patient feature matrices what derived form
\texttt{word2vec} embedding.

We verify the cohort discover studies by evaluating the clustering using three
popular criteria: $\mathit{Rand\ index}$, $Purity$ and
$\mathit{normalized \ mutual \ information (NMI)}$.

$\mathit{Rand\ index}$ is frequently used in data clustering, it is computed as
following in \cite{rand1971objective}:
$$RI = \left(TP+TN\right)/\binom{n}{2}$$ where $TP$ counts the number of right
decision we have made on grouping pairs of patients who are in the same cohort
into one cluster, $TN$ is the number of pairs of patients who came from
different cohorts are grouped into dissimilar categories. In general, bad
clustering have $RI$ values close to 0, a perfect clustering has a $RI$ of 1.

$Purity$ is one of very primary validation measure to evaluate the cluster
quality. We compute $Purity$ as defined in \cite{manning2008introduction}:
$$Purity\left( Cluster,Cohort \right) = \frac{1}{N}\sum_I\max_j|p_i \cap q_j|$$
where $Cluster=\{p_1,p_2,...,p_I\}$ is the set of clusters,
$Cohort=\{q_1,q_2,...q_J\}$ is the group of classes, or cohorts in our case. The
cohort is identified by the categories of dominant patients in cluster. Similar
to $Rand\ index$, the $Purity$ has upper bound of 1 corresponding to the perfect
match between the partitions and lower bound of 0 that indicates the opposite.

$\mathit{NMI}$ measures the information shared by the two clusters,thus can
be adopted as a clustering similarity measure. We follow the form defined in
\cite{meilua2007comparing} to calculate $\mathit{NMI}$ value.
$$\mathit{NMI}\left(C_{luster},C_{ohort}\right)=
\frac{\mathit{I}\left(C_{luster},C_{ohort}\right)}
{\left[\mathit{H}\left(C_{luster}\right)+\mathit{H}\left(C_{ohort}\right)\right]
  /2}$$ where,
$$\mathit{I}\left(X,Y\right)=\sum_{x \in  X}\sum_{y \in Y}p\left(x,y\right)\log
\frac{p\left(x,y\right)}{p\left(x\right)p\left(y\right)}$$ is the mutual
information between the random variables $X$ and $Y$,
$$\mathit{H}\left(X\right)= -\sum_{x \in X}p\left(x\right)\log p\left(x\right)$$
is the information entropy of a discrete random
variable$X$. $p\left(x\right),p\left(x,y\right)$ are the probabilities of a
object being in cluster $X$ and in the the intersection of $X$ and $Y$. The
$\mathit{NMI}$ has a fixed lower bound of 0 and upper bound of 1. In our case,
$\mathit{NMI}\left(C_{luster},C_{ohort}\right)$ takes its maximum value of 1
when grouping clusters are identical to the real cohorts, if the partition
found is totally independent of the real cohorts, then
$\mathit{NMI}\left(C_{luster},C_{ohort}\right) = 0$.

There are other popular measures for cluster evaluation, such as $Precision$
\cite{zhao2001criterion}, $Recall$, and their harmonic mean, $F-measure$
\cite{van1974foundation}. We also present our results by these evaluation.

\subsection{Results and discussion}

\begin{figure}[!htbp]
  \centering
  \subfloat[]{
    \label{fig:comparison-ri}
    \includegraphics[width=0.37\textwidth]{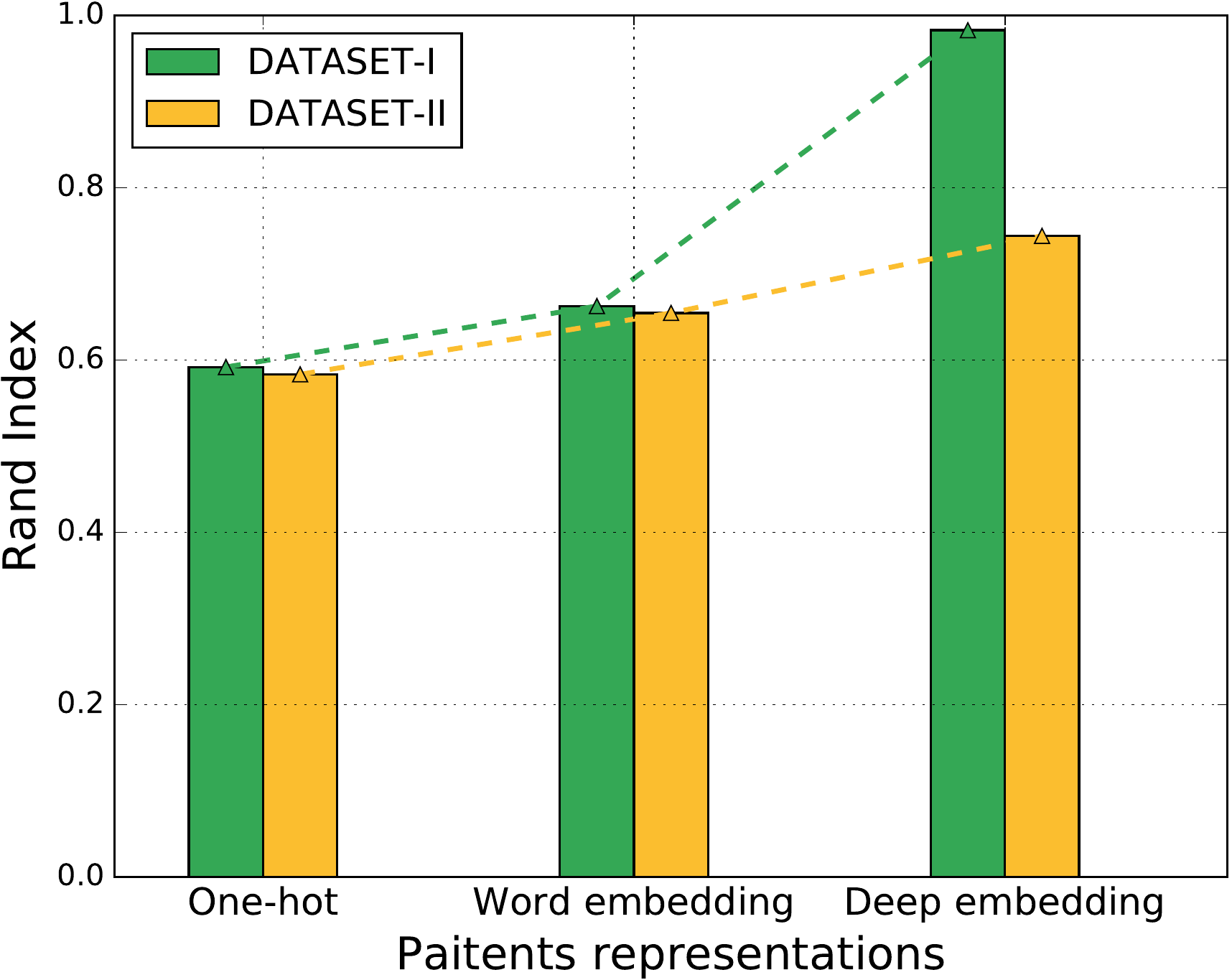}
  }
  \\
  \subfloat[]{
    \label{fig:comparison-purity}
    \includegraphics[width=0.37\textwidth]{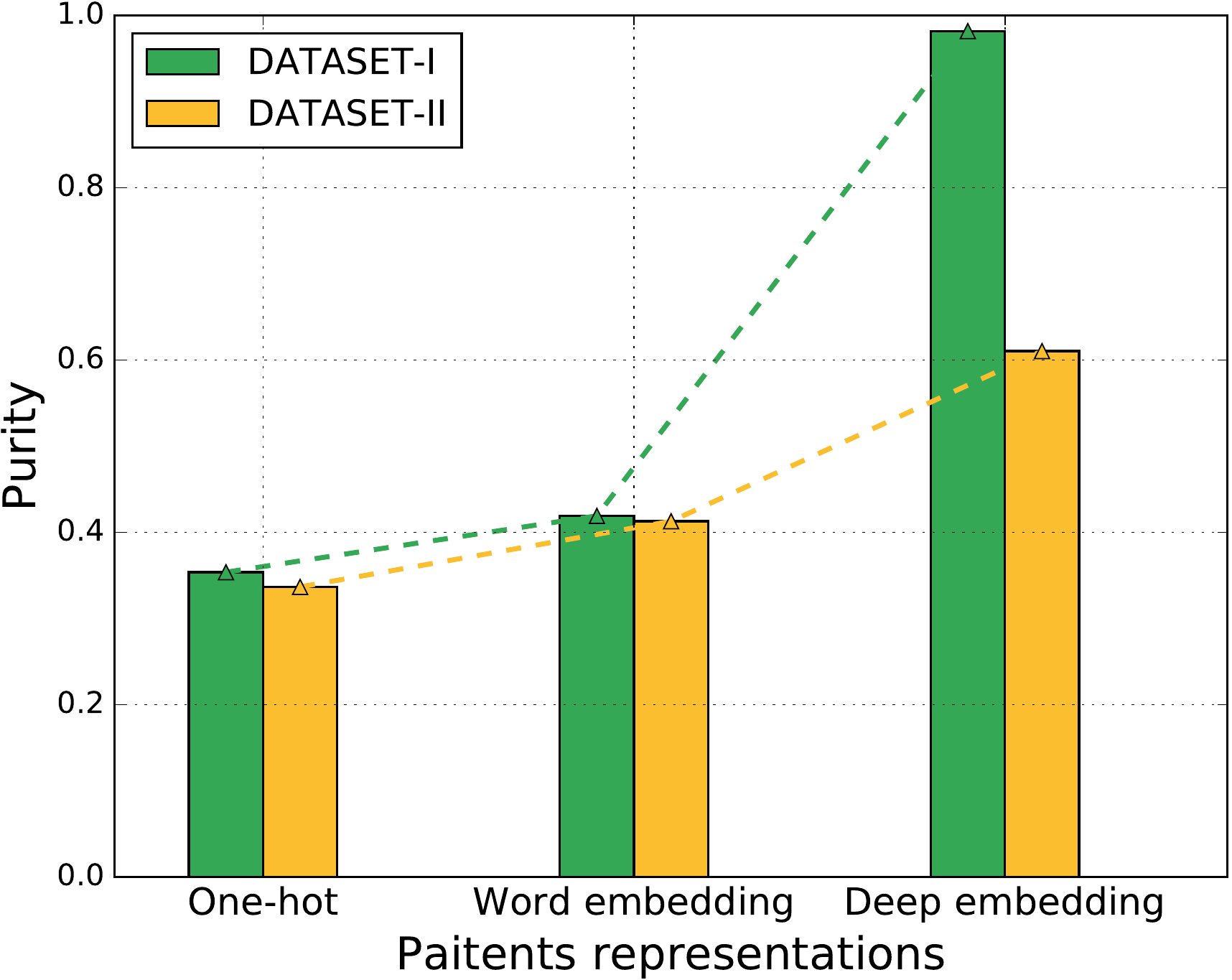}
  }
  \\
  \subfloat[]{
    \label{fig:comparison-nmi}
    \includegraphics[width=0.37\textwidth]{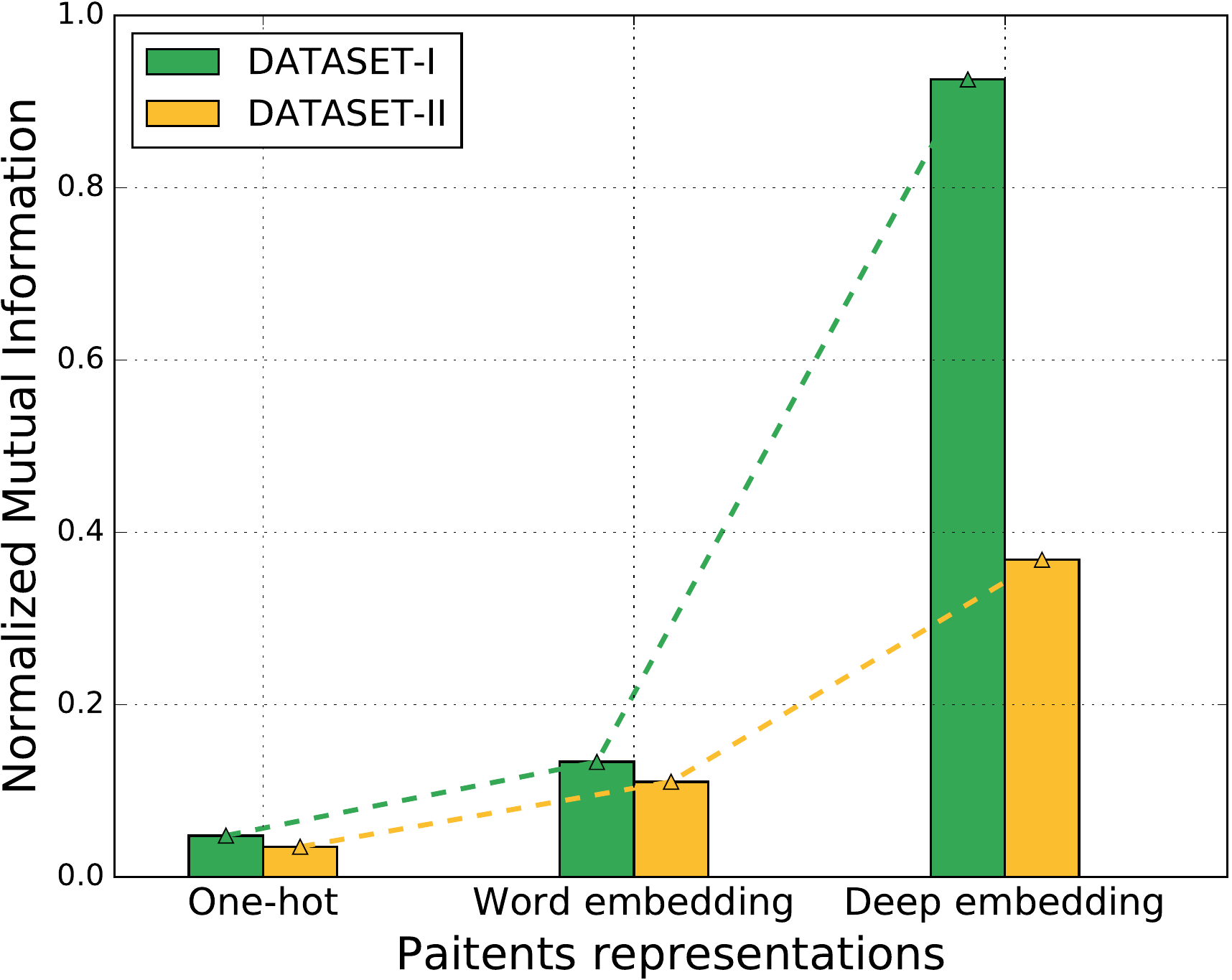}
  }
  \caption{Performance of different representations.}
  \label{fig:represent-comparison}
\end{figure}

\begin{figure*}[!htbp]
  \centering
  \subfloat[]{
    \label{fig:word2vec_ri}
    \includegraphics*[width=0.3\textwidth]{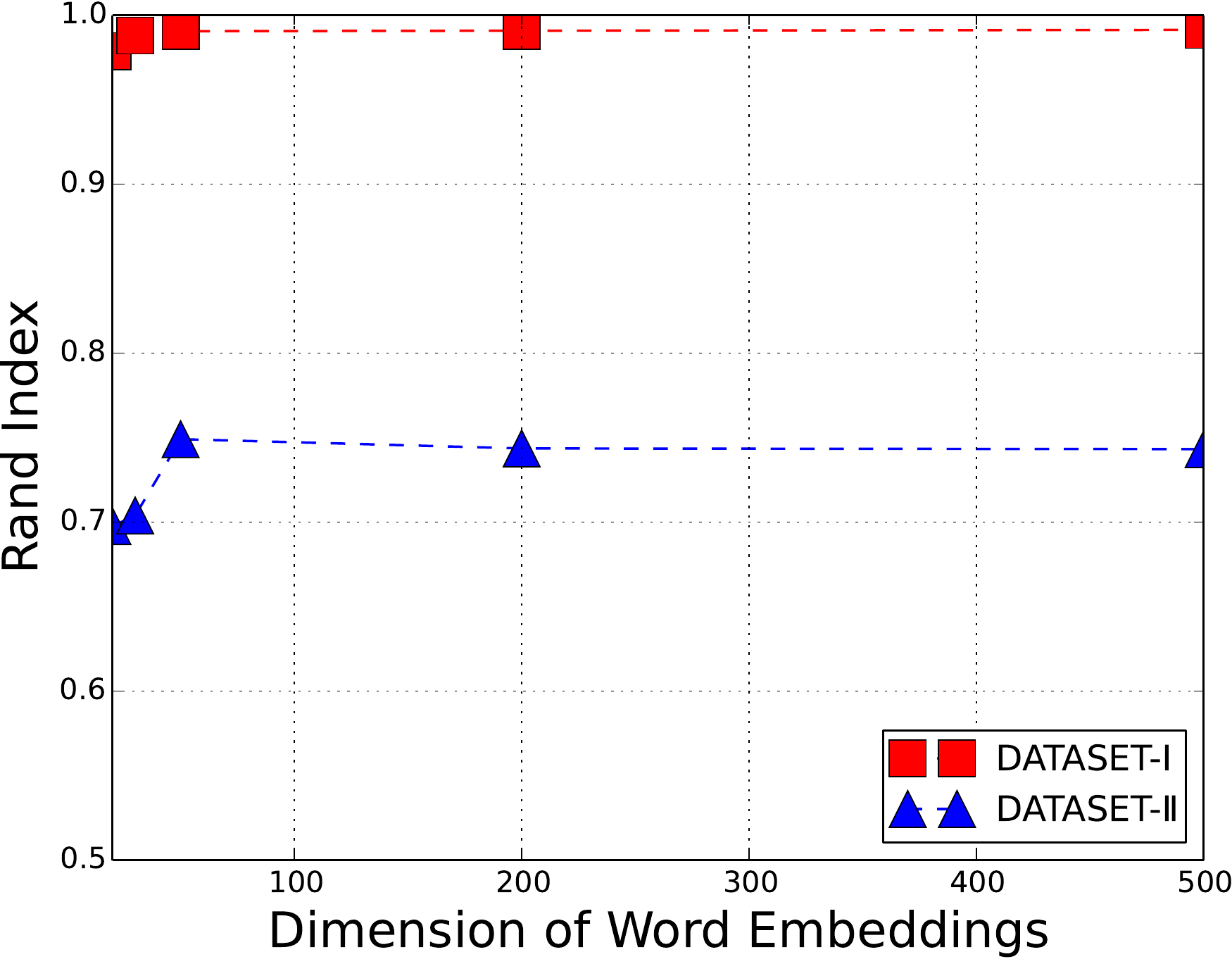}
  }
  \subfloat[]{
    \label{fig:word2vec_purity}
    \includegraphics*[width=0.3\textwidth]{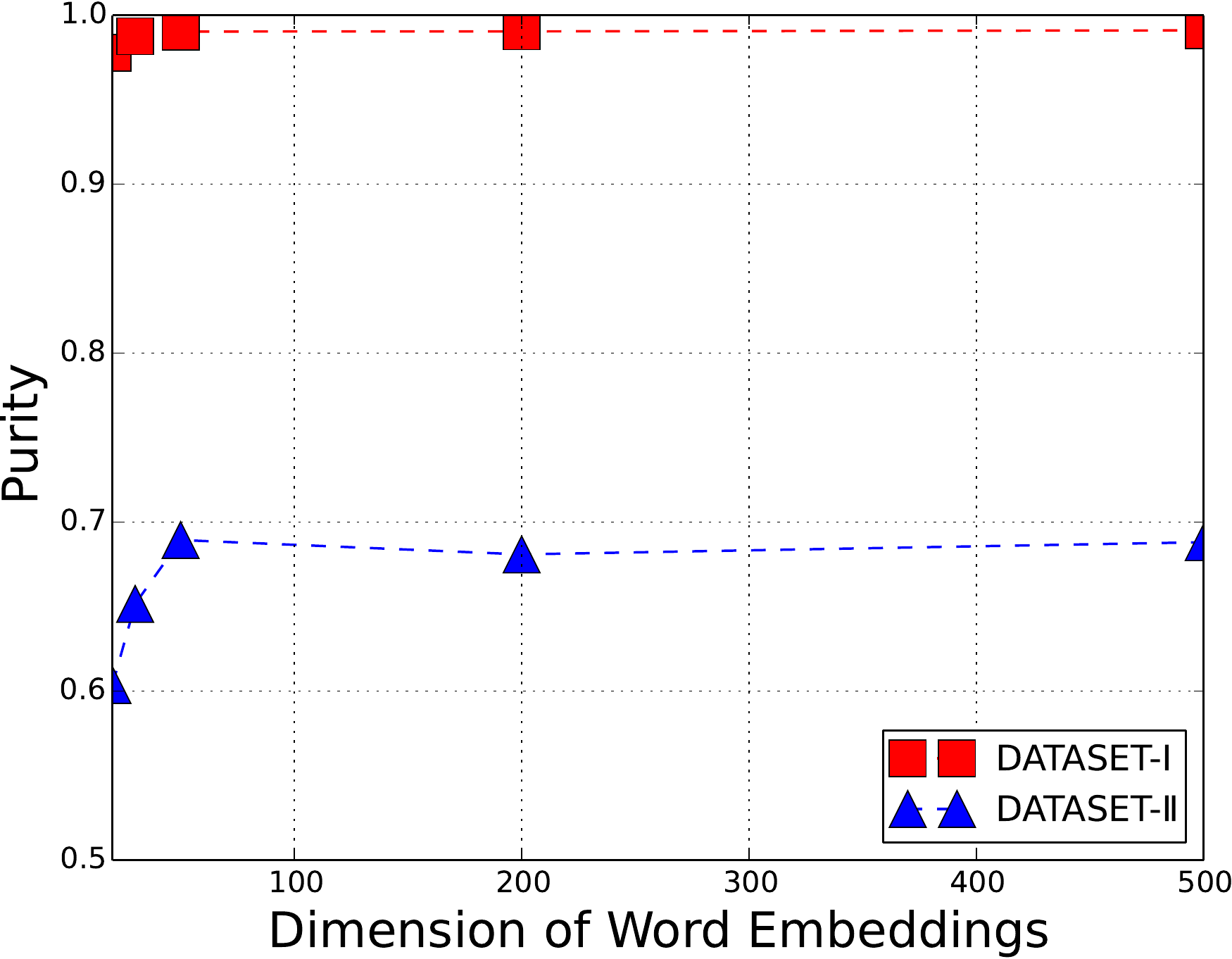}
  }
  \subfloat[]{
    \label{fig:word2vec_nmi}
    \includegraphics*[width=0.3\textwidth]{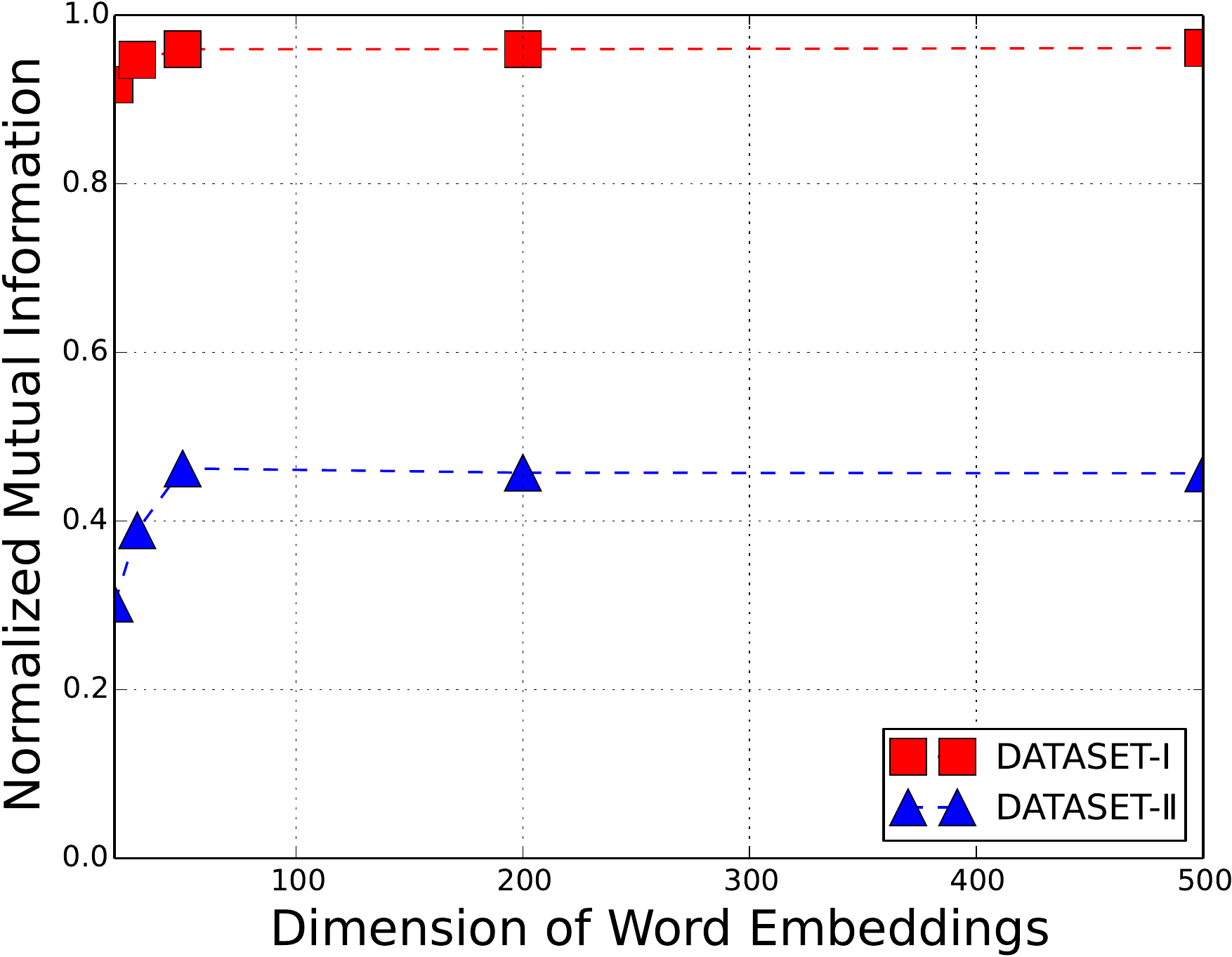}
  }
  \\
  \subfloat[]{
    \label{fig:numfeature_ri}
    \includegraphics*[width=0.3\textwidth]{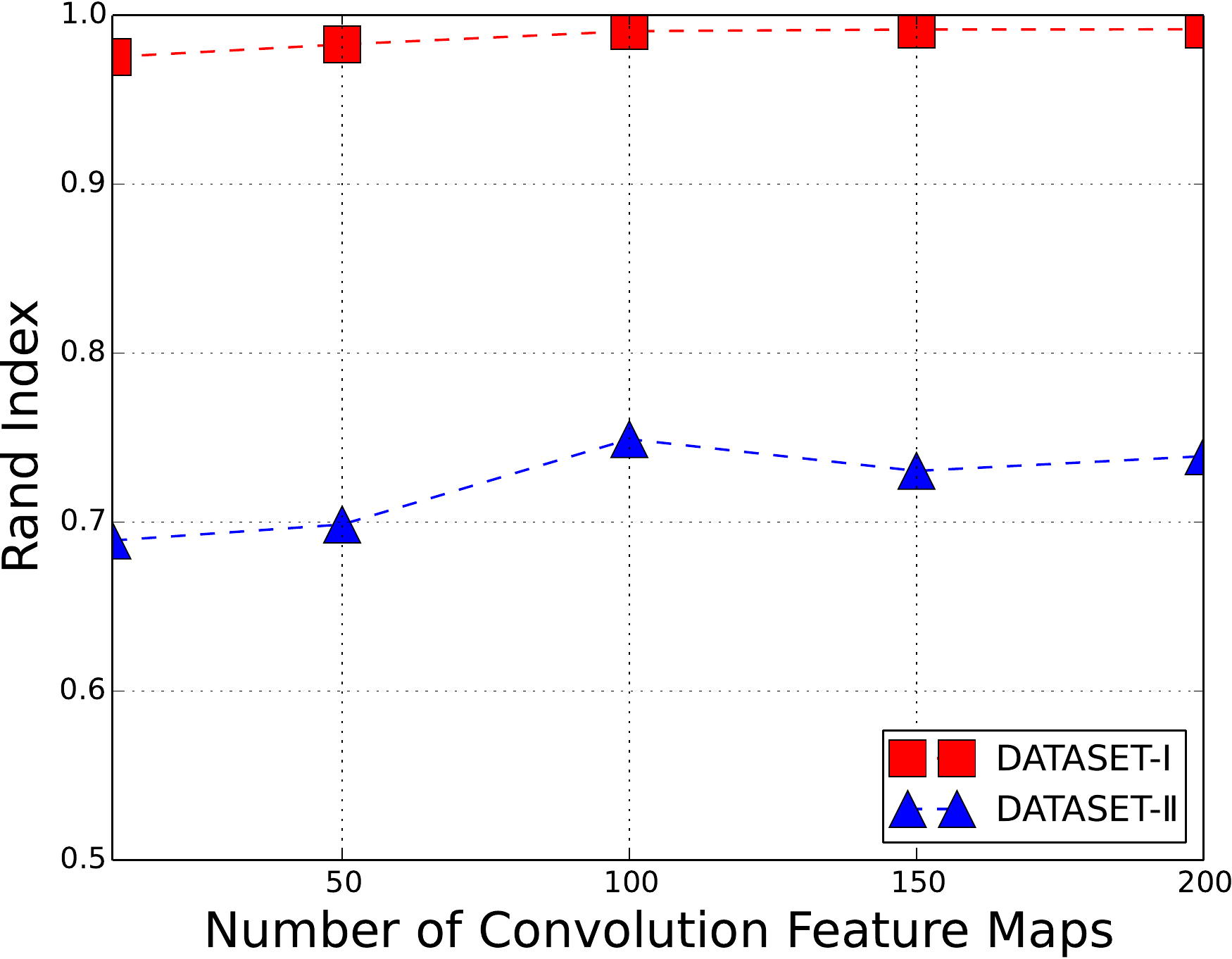}
  }
  \subfloat[]{
    \label{fig:numfeature_purity}
    \includegraphics*[width=0.3\textwidth]{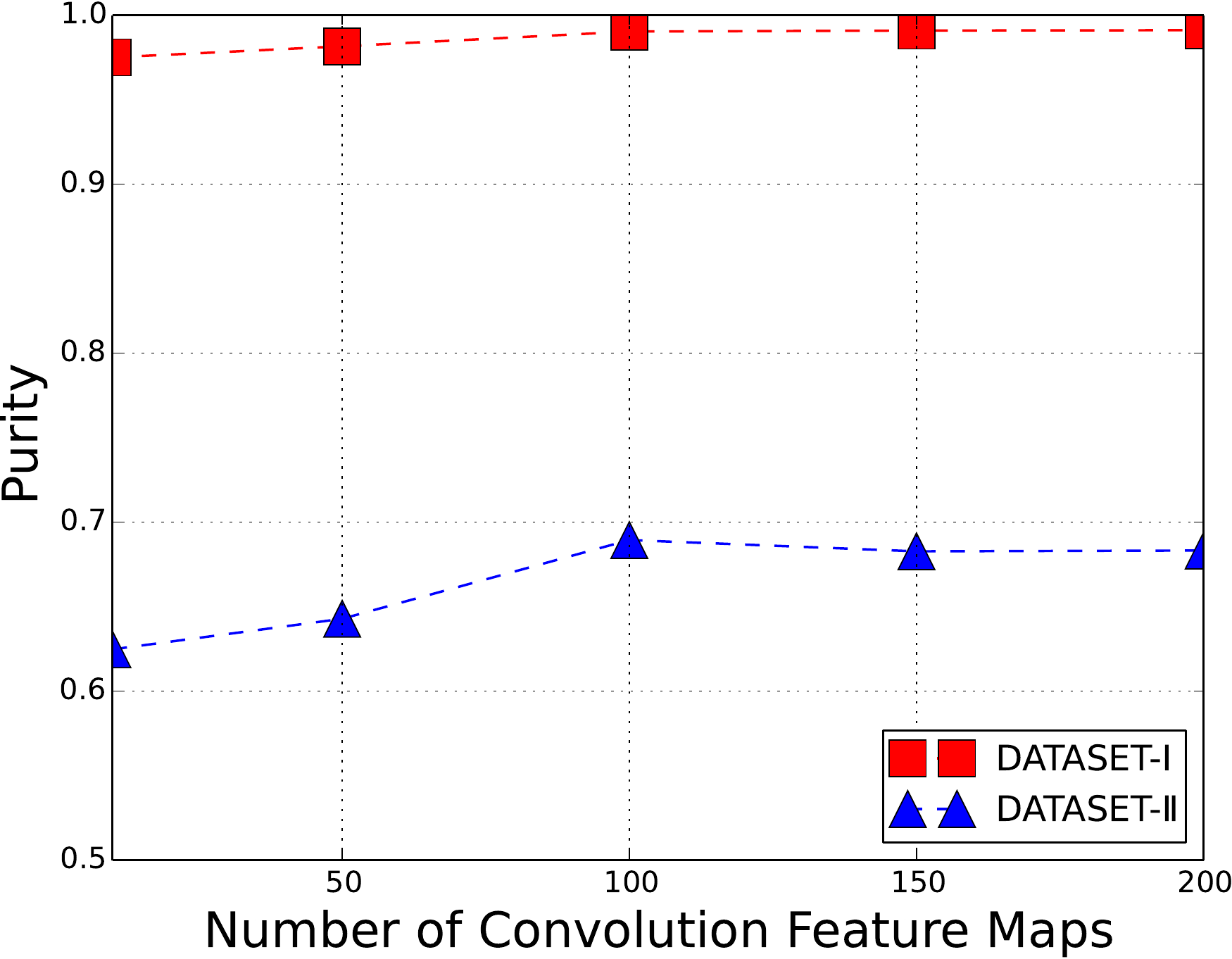}
  }
  \subfloat[]{
    \label{fig:numfeature_nmi}
    \includegraphics*[width=0.3\textwidth]{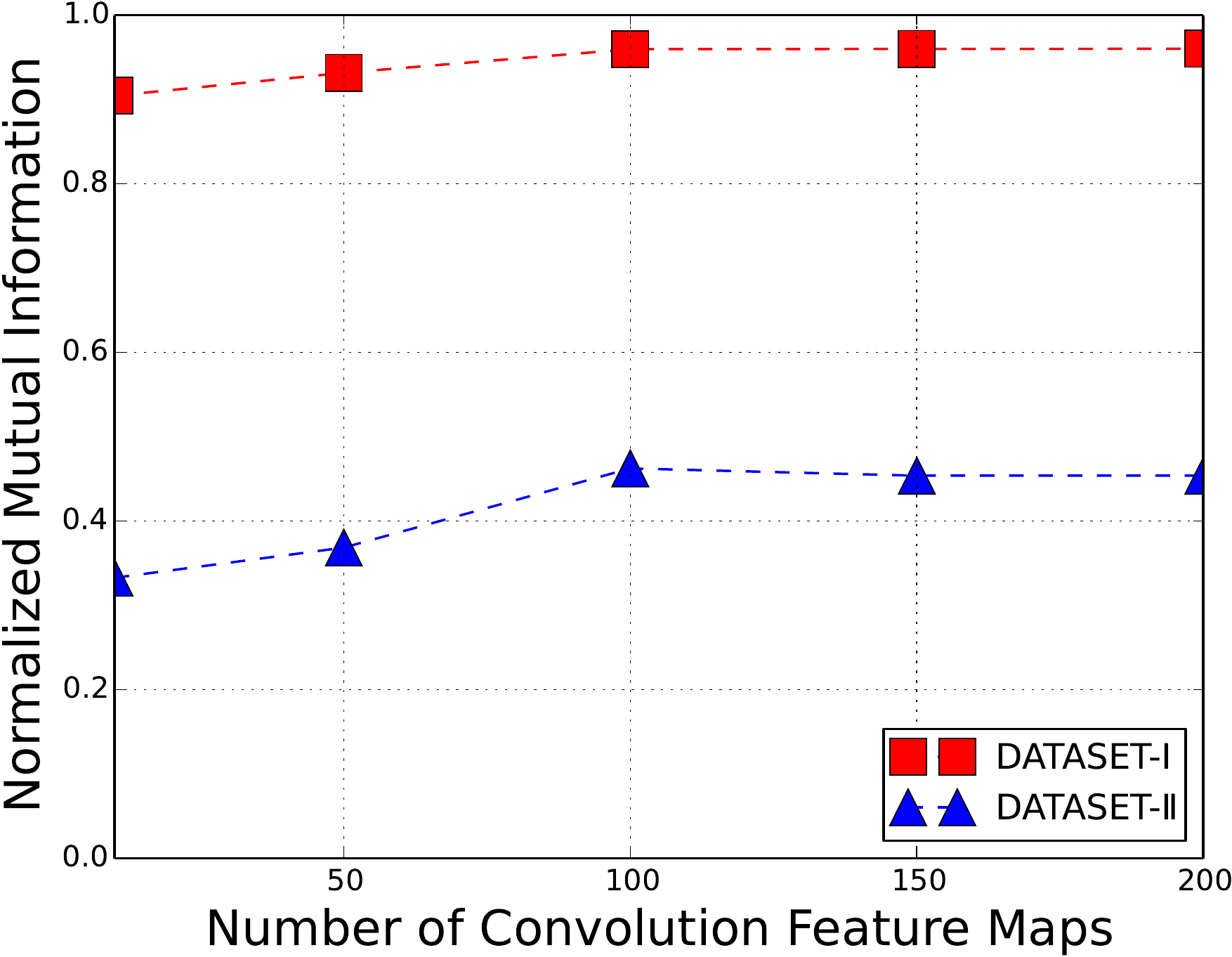}
  }
  \\
  \subfloat[]{
    \label{fig:filterwidth_ri}
    \includegraphics*[width=0.3\textwidth]{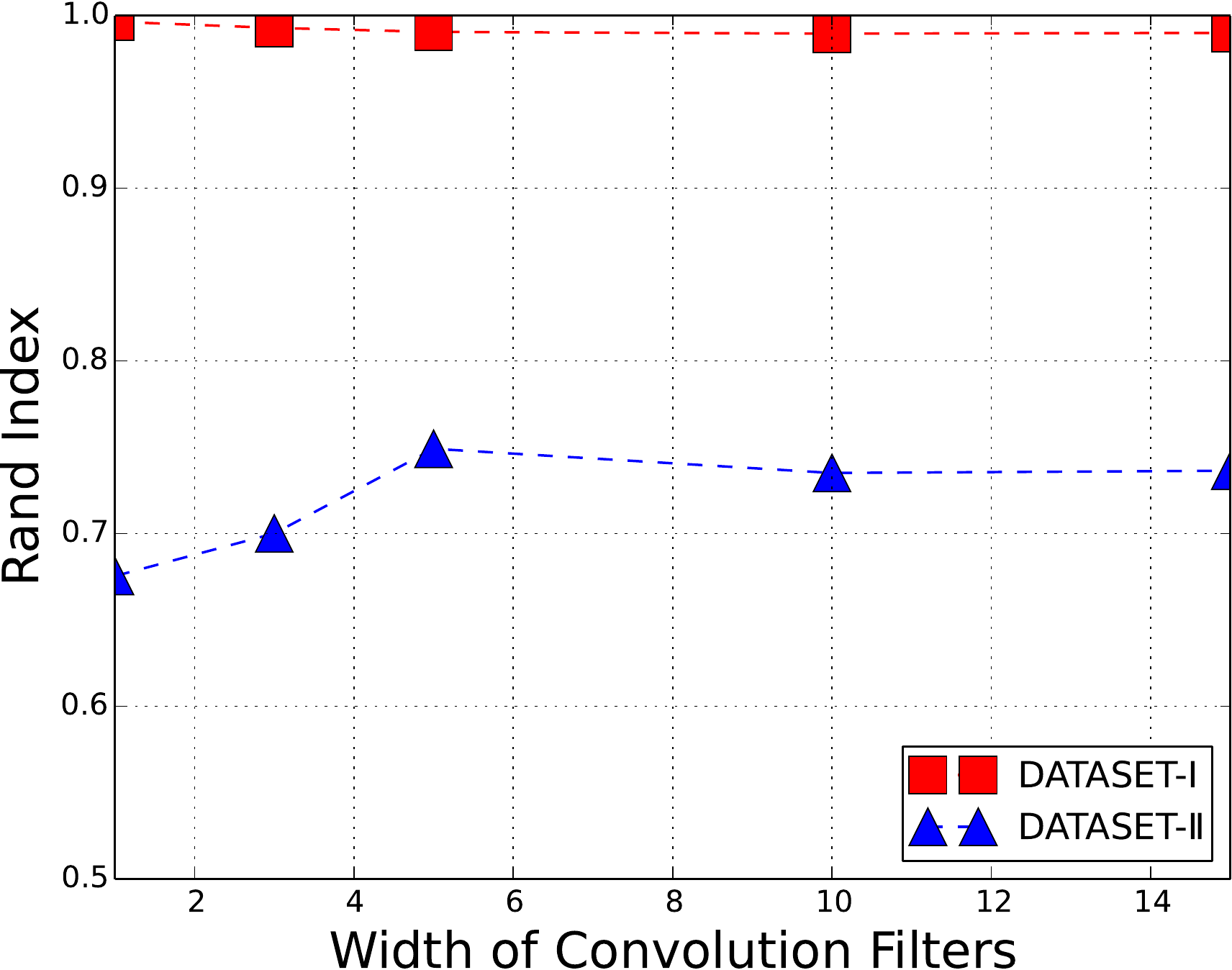}
  }
  \subfloat[]{
    \label{fig:filterwidth_purity}
    \includegraphics*[width=0.3\textwidth]{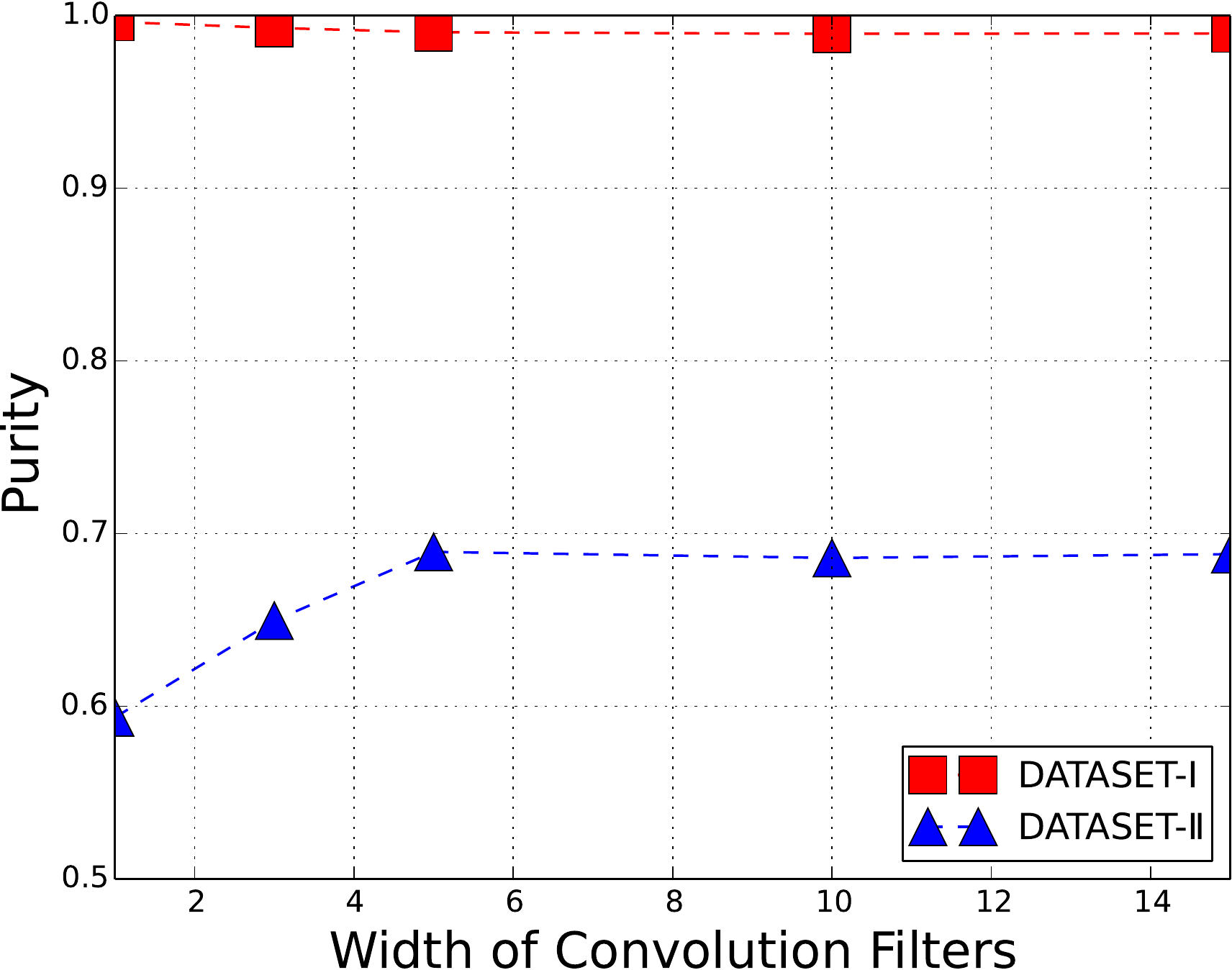}
  }
  \subfloat[]{
    \label{fig:filterwidth_nmi}
    \includegraphics*[width=0.3\textwidth]{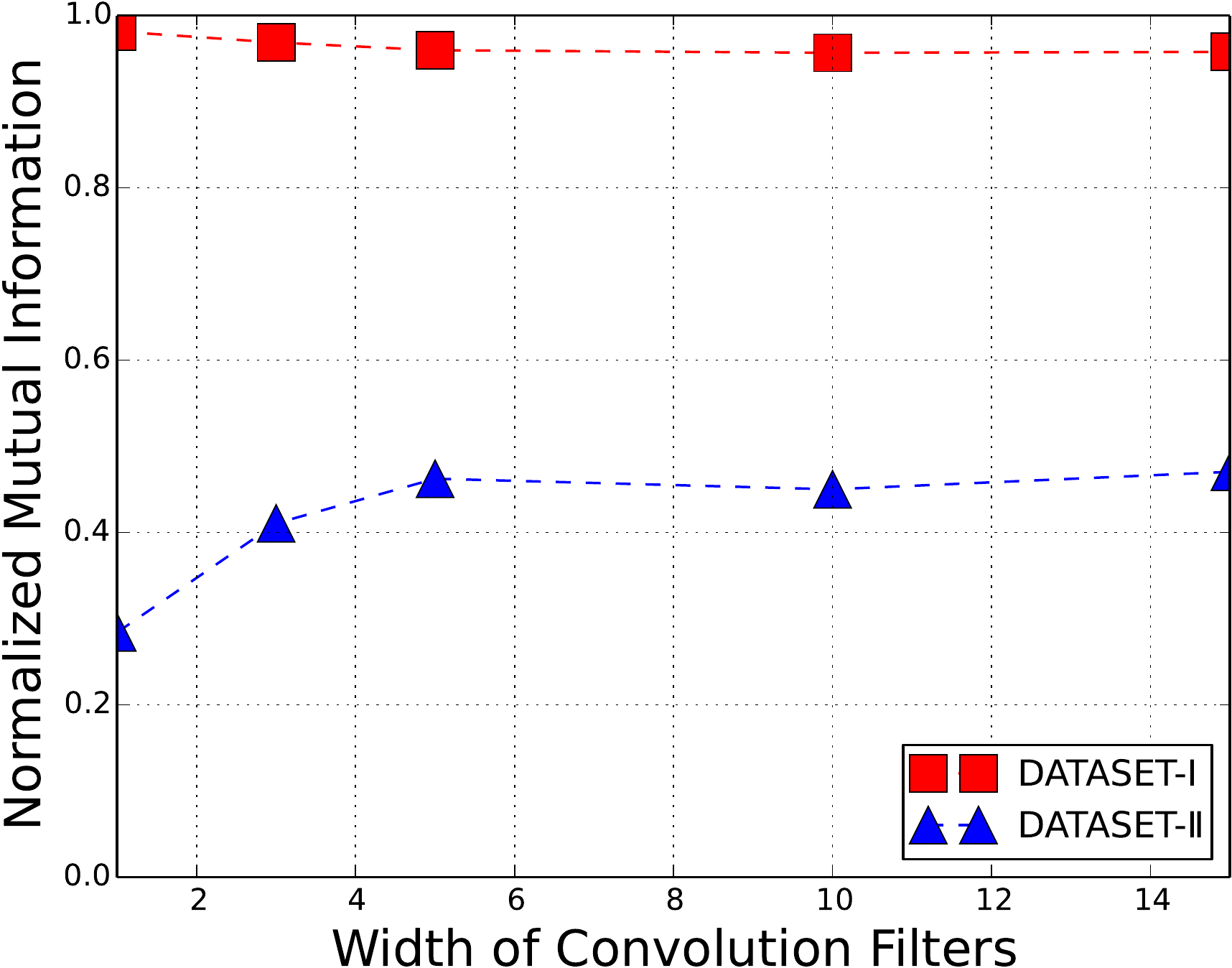}
  }
  \vspace{-0.1in}
  \caption{Results on the hyper-parameters
    optimizations. \protect\subref{fig:word2vec_ri},
    \protect\subref{fig:word2vec_purity}, \protect\subref{fig:word2vec_nmi}
    together varify the effect of word embedding dimension on clustering
    performance. \protect\subref{fig:numfeature_ri},
    \protect\subref{fig:numfeature_purity}, \protect\subref{fig:numfeature_nmi}
    measures the efficacy of variation on convolution feature
    maps. \protect\subref{fig:filterwidth_ri},
    \protect\subref{fig:filterwidth_purity},
    \protect\subref{fig:filterwidth_nmi} make determinations of convolution
    feature filters width.}
  \label{fig:hyperparameters-opt}
\end{figure*}

\begin{table*}[!htbp]
  \centering
  \begin{tabular}{lccccccc}
    \toprule
    &\multicolumn{3}{c}{DATASET-\Rmnum{1}}&& \multicolumn{3}{c}{DATASET-\Rmnum{2}} \\
    Method& $RI$& $Purity$& $NMI$&& $RI$& $Purity$& $NMI$ \\
    \midrule
    $\mathtt{KMeans^1}$& 0.5919&0.3538& 0.0481&& 0.5834& 0.3367& 0.0351 \\
    $\mathtt{Active \ PCKMeans^1}$& 0.6506& 0.4801& 0.0976&& 0.6451& 0.4410& 0.0682 \\
    $\mathtt{KMeans^2}$& 0.6627& 0.4192& 0.1336&& 0.6547& 0.4129& 0.1106 \\
    $\mathtt{Active \ PCKMeans^2}$& 0.6796& 0.5610& 0.1682&& 0.6794& 0.5487& 0.1622 \\
    $\mathtt{Matric \ Learning}$ &0.6732 &0.5103 &0.1049 && 0.6475& 0.4013 & 0.1055 \\
    \midrule
    Our model($RV$)& 0.6679& 0.4457& 0.2301&& 0.6285& 0.3858& 0.1037 \\
    Our model($dCov$)&0.6708& 0.4475& 0.2361&& 0.6268& 0.3858& 0.1048 \\
    Our model(CNN)& \textbf{0.9887}& \textbf{0.9882}& \textbf{0.9516}&& \textbf{0.7491}& \textbf{0.6894}& \textbf{0.4624} \\
    \bottomrule
  \end{tabular}
  \caption{Evaluations of cohort discovering on DATASET-\Rmnum{1},
    DATASET-\Rmnum{2}. $\mathtt{KMeans^1}$, $\mathtt{Active \ PCKMeans^1}$
    groups patients with One-hot representations, where $\mathtt{KMeans^2}$,
    $\mathtt{Active \ PCKMeans^2}$ adopt the Shallow embeddings to match similar
    patients pairs. We contrast the performance of our model at
    DATASET-\Rmnum{1}, DATASET-\Rmnum{2} with the same parameters. Values of
    $RI$,$Purity$,$NMI$ presented in the table are average of a group of
    results.}
  \label{tab:results-cohorts}
\end{table*}

\begin{figure*}[!htbp]
    \centering
    \includegraphics[width = 0.58\textwidth]{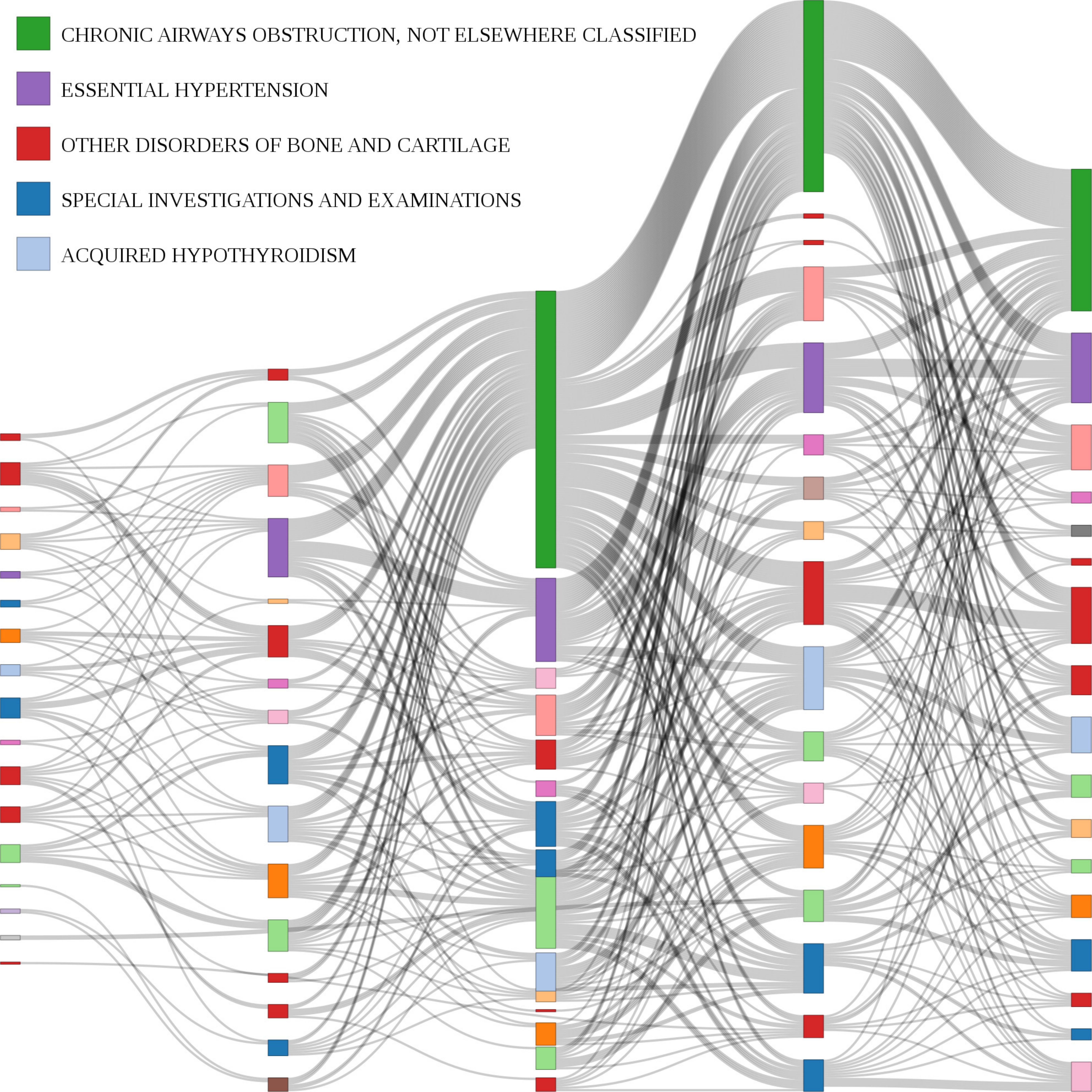}
    \caption{An Example of sankey-pathway on the COPD dataset. Each color represents different features and only top frequent feature names are listed.}
    \label{fig:sankey-pathway}
  \end{figure*}

\subsubsection{Performance Comparison}
\label{sec:Performance-comparison}
Table \ref{tab:results-cohorts} summaries the results of clustering. As we can see, the deep model with feature embedding is clearly superior to others.
On DATASET-\Rmnum{1}, the deep embedding model
achieves an average $Rand\ index$ of $0.9887$, comparing with the second best one with $0.6796$. Measured by $Purity$ and $NMI$, it can achieve the performances of $0.9882$ and $0.9516$, separately, which also outperforms others with a margin. The superiority of the model is illustrated in
DATASET-\Rmnum{2} as well, which is a more difficult task. Measured by  $Purity$ and $NMI$, $\mathtt{KMeans^1}$ and $\mathtt{Active\ PCKMeans^1}$ achieve 0.3367, 0.0351 and 0.4410, 0.0682 separately. $\mathtt{KMeans^2}$ and $\mathtt{Active\ PCKMeans^2}$ can only improve 11\% and 25\% on $Purity$ respectively. On the other hand, our CNN model achieves about more than 50\% improvement over them.

As a reasonal explanation, we view that the deep features learning can be
viewed as a two-stage model. During the first stage, the clinical features of
each patients are summarized in the shallow \texttt{word2vec} embedding, making
progress with nearly $10\%$ improvement. Next, global features are learned base
on local context features came from \texttt{word2vec}. The deep learning
representation makes continuous improvement, which leads to a ultimate
expression of patients. Figure \ref{fig:represent-comparison} shows how
expressive representations of patients contribute to match patients cohorts.
With a significant $48\%$ improvement produced in our experiments, we
demonstrate the effectiveness of our deep embedding model in expressively representing patients.

\subsubsection{Parameters Optimization}
\label{sec:model-selection}

Figure \ref{fig:hyperparameters-opt} illustrates the optimizations of
hyper-parameters in our model. The line charts in one row assess what effects the
variation has on grouping patients. As results summarized in Figure
\ref{fig:word2vec_ri}, Figure \ref{fig:word2vec_purity}, Figure
\ref{fig:word2vec_nmi}, the dimension of medical event embedding have little
effect on DATASET-\Rmnum{1}. That because our deep learning model have
successfully obtained the primary features in the patients representations,
achieving nearly perfect 1 of $RI$, $Purity$, and $NMI$. We make determinations
based on DATASET-\Rmnum{2}. According to the performance lines shown in the
figures, three clustering evaluations we choose---$RI$, $Purity$ and $NMI$
achieves the best performance at the same time, with 50 dimensionality
embedding, 100 feature maps, and 5 convolution filter width. The consistent
performances of different measures assessed in our experiments convince us that
the optimizations of parameters are correct and free of bias.

\subsubsection{Discussions}
\label{sec:discussions}

Table \ref{tab:results-cohorts} provides the comparisons of clustering results
on DATASET-\Rmnum{1} and DATASET-\Rmnum{2}. The results of deep embedding on
DATASET-\Rmnum{2} exhibit a steady outperformance over other methods. On
DATASET-\Rmnum{2}, the deep embedding model is trained with fewer medical events
than DATASET-\Rmnum{1}. As expected, the evaluations of identifying patient
cohorts is slight affected by the data we deal with. Compared to the sterling
performances on DATASET-\Rmnum{1}, the $RI$ resulted in DATASET-\Rmnum{2} drops
to $0.75$ with loss of $0.23$. One simple but reasonable explanation is that the
events removing from dataset cause a loss of many a medical features. Even
though, our deep model extract remain features effectively offering promising
performance. To sum up, we verify that our deep learning method works
effectively in representing patients and learning specific features that are not
present or missing.

Table \ref{tab:results-cohorts} also reports the comparisons on effectiveness of
our supervised and unsupervised measurement of patient similarity. On average,
the unsupervise measurements---$RV$ and $dCov$, respectively gets $RI$ of $0.67$
and $NMI$ of $0.23$, which are 31\% and 72\% lower than the deep learning
model($0.97$). Although, it's worthwhile to mention that our unsupervise models
achieve at least 12\% improvement over the baseline. Comparing to the
semi-supervise method $\mathtt{Active \ PCKMeans}$ proposed in
\cite{basu2004active}, our model achieves the same performance but do not need
training examples. These comparisons suggest that our models consistently and
significantly surpass other patient representations.

\subsubsection{Visual Analysis}
\label{sec:visual}
As we have achieved a definitely accurate measure of patient similarity, we make
a study on medical events sequence mining for representation. In order to
discovery the medical pattern hidden behind the EHR about COPD, we select
top-100 similar patients from the COPD cohort , whom are grouped into true
cluster by our method. We extract the common events occurred in the EMR of many
patients.The SanKey diagram presents the progression path of the medical events
collected from patients EHR. As shown in the Figure, the green,purple and red
bar are related closely, which respectively represent Chronic Airways
Obstruction, Essential Hypertension, Other Disorder of Bone and Cartilage. The
interactions of those diseases presented in the diagram has been validated by
lots of medical research in the real world, that convinces us of the
applicability of our model.

In summary, the results of experiments clearly demonstrate the effectiveness of
the deep model with medical feature embedding on real EHR data.  Theoretically,
our model benefits from the large number of convolutional filters and lower
event embedding dimensionality. It is notable that our model has several
important hyper-parameters like \texttt{word2vec} window size, dimensionality of
clinical event vector, the number of convolutional filters. Selecting a set of
optimal parameters settings can also bring the benefit of the performance. To be
more realistic, we narrow down the scopes of variations and select the best
performance values.
\section{Conclusions}
\label{sec:conclusions}

Patient similarity assessment is the enabling technique for various healthcare
applications, such as disease sub-typing and evidence based medicine. However,
due to the complexity of medical data, extracting effective patient
representations confronts distinct challenges. Though useful, most existing
models proposed to discover hidden patterns in EHRs overlook the temporal
information of medical events. In this paper, we propose a deep learning
framework to learn patient representations for similarity measuring, in which
the temporal properties of EHRs are preserved. The experimental results show
that our model achieves significantly better representations over the baselines,
which enables more accurate patient cohort discovery. Our next plans include
solving the data irregularity issue by adding the time interval information and
applying this techniques in other domain, such as health visualization. Besides,
it can be observed in the experiments that our unsupervised scheme also
succeeded in matching similar patients.

\section{Acknowledgement}

This work is sponsored by ``The Fundamental Theory and Applications of Big Data with Knowledge Engineering" under the National Key Research and Development Program of China with grant number 2016YFB1000903; National Science Foundati on
of China under Grant Nos. 61428206; Ministry of Education Innovation Research Team No. IRT13035.

\bibliography{reference} \bibliographystyle{IEEEtran}

\begin{thebibliography}{10}
\providecommand{\url}[1]{#1}
\csname url@samestyle\endcsname
\providecommand{\newblock}{\relax}
\providecommand{\bibinfo}[2]{#2}
\providecommand{\BIBentrySTDinterwordspacing}{\spaceskip=0pt\relax}
\providecommand{\BIBentryALTinterwordstretchfactor}{4}
\providecommand{\BIBentryALTinterwordspacing}{\spaceskip=\fontdimen2\font plus
\BIBentryALTinterwordstretchfactor\fontdimen3\font minus
  \fontdimen4\font\relax}
\providecommand{\BIBforeignlanguage}[2]{{%
\expandafter\ifx\csname l@#1\endcsname\relax
\typeout{** WARNING: IEEEtran.bst: No hyphenation pattern has been}%
\typeout{** loaded for the language `#1'. Using the pattern for}%
\typeout{** the default language instead.}%
\else
\language=\csname l@#1\endcsname
\fi
#2}}
\providecommand{\BIBdecl}{\relax}
\BIBdecl

\bibitem{Cheng2016Risk}
Y.~Cheng, F.~Wang, P.~Zhang, and J.~Hu, ``Risk prediction with electronic
  health records: A deep learning approach,'' 2016.

\bibitem{sun2012supervised}
J.~Sun, F.~Wang, J.~Hu, and S.~Edabollahi, ``Supervised patient similarity
  measure of heterogeneous patient records,'' \emph{ACM SIGKDD Explorations
  Newsletter}, vol.~14, no.~1, pp. 16--24, 2012.

\bibitem{mikolov2013distributed}
T.~Mikolov, I.~Sutskever, K.~Chen, G.~S. Corrado, and J.~Dean, ``Distributed
  representations of words and phrases and their compositionality,'' in
  \emph{Advances in neural information processing systems}, 2013, pp.
  3111--3119.

\bibitem{mikolov2013efficient}
T.~Mikolov, K.~Chen, G.~Corrado, and J.~Dean, ``Efficient estimation of word
  representations in vector space,'' \emph{arXiv preprint arXiv:1301.3781},
  2013.

\bibitem{josse2013measures}
J.~Josse and S.~Holmes, ``Measures of dependence between random vectors and
  tests of independence. literature review,'' \emph{arXiv preprint
  arXiv:1307.7383}, 2013.

\bibitem{szekely2007measuring}
G.~J. Sz{\'e}kely, M.~L. Rizzo, N.~K. Bakirov \emph{et~al.}, ``Measuring and
  testing dependence by correlation of distances,'' \emph{The Annals of
  Statistics}, vol.~35, no.~6, pp. 2769--2794, 2007.

\bibitem{chan2010machine}
L.~Chan, T.~Chan, L.~Cheng, and W.~Mak, ``Machine learning of patient
  similarity: A case study on predicting survival in cancer patient after
  locoregional chemotherapy,'' in \emph{Bioinformatics and Biomedicine
  Workshops (BIBMW), 2010 IEEE International Conference on}.\hskip 1em plus
  0.5em minus 0.4em\relax IEEE, 2010, pp. 467--470.

\bibitem{wang2012medical}
F.~Wang, J.~Hu, and J.~Sun, ``Medical prognosis based on patient similarity and
  expert feedback,'' in \emph{Pattern Recognition (ICPR), 2012 21st
  International Conference on}.\hskip 1em plus 0.5em minus 0.4em\relax IEEE,
  2012, pp. 1799--1802.

\bibitem{wang2012towards}
F.~Wang, N.~Lee, J.~Hu, J.~Sun, and S.~Ebadollahi, ``Towards heterogeneous
  temporal clinical event pattern discovery: a convolutional approach,'' in
  \emph{Proceedings of the 18th ACM SIGKDD international conference on
  Knowledge discovery and data mining}.\hskip 1em plus 0.5em minus 0.4em\relax
  ACM, 2012, pp. 453--461.

\bibitem{chenips}
Z.~Che, Y.~Cheng, Z.~Sun, and Y.~Liu, ``Exploiting convolutional neural network
  for risk prediction with medical feature embedding,'' \emph{CoRR}, vol.
  abs/1701.07474, 2017.

\bibitem{ng2015personalized}
K.~Ng, J.~Sun, J.~Hu, and F.~Wang, ``Personalized predictive modeling and risk
  factor identification using patient similarity,'' \emph{AMIA Summits on
  Translational Science Proceedings}, vol. 2015, p. 132, 2015.

\bibitem{kasabov2010integrated}
N.~Kasabov and Y.~Hu, ``Integrated optimisation method for personalised
  modelling and case studies for medical decision support,''
  \emph{International Journal of Functional Informatics and Personalised
  Medicine}, vol.~3, no.~3, pp. 236--256, 2010.

\bibitem{sewitch2004clustering}
M.~J. Sewitch, K.~Leffondr{\'e}, and P.~L. Dobkin, ``Clustering patients
  according to health perceptions: relationships to psychosocial
  characteristics and medication nonadherence,'' \emph{Journal of psychosomatic
  research}, vol.~56, no.~3, pp. 323--332, 2004.

\bibitem{henao2013patient}
R.~Henao, J.~Murray, G.~Ginsburg, L.~Carin, and J.~E. Lucas, ``Patient
  clustering with uncoded text in electronic medical records,'' in \emph{AMIA
  Annual Symposium Proceedings}, vol. 2013.\hskip 1em plus 0.5em minus
  0.4em\relax American Medical Informatics Association, 2013, p. 592.

\bibitem{huang2013spectral}
G.~T. Huang, K.~I. Cunningham, P.~V. Benos, and C.~S. CHENNUBHOTLA, ``Spectral
  clustering strategies for heterogeneous disease expression data,'' in
  \emph{Pacific Symposium on Biocomputing. Pacific Symposium on
  Biocomputing}.\hskip 1em plus 0.5em minus 0.4em\relax NIH Public Access,
  2013, p. 212.

\bibitem{kiela2014learning}
D.~Kiela and L.~Bottou, ``Learning image embeddings using convolutional neural
  networks for improved multi-modal semantics.'' in \emph{EMNLP}.\hskip 1em
  plus 0.5em minus 0.4em\relax Citeseer, 2014, pp. 36--45.

\bibitem{severyn2015learning}
A.~Severyn and A.~Moschitti, ``Learning to rank short text pairs with
  convolutional deep neural networks,'' in \emph{Proceedings of the 38th
  International ACM SIGIR Conference on Research and Development in Information
  Retrieval}.\hskip 1em plus 0.5em minus 0.4em\relax ACM, 2015, pp. 373--382.

\bibitem{10.1093/jamia/ocx090}
Y.~Luo, Y.~Cheng, z.~Uzuner, P.~Szolovits, and J.~Starren, ``{Segment
  convolutional neural networks (Seg-CNNs) for classifying relations in
  clinical notes},'' \emph{Journal of the American Medical Informatics
  Association}, vol.~25, no.~1, pp. 93--98, 2017.

\bibitem{ehr-gan}
Z.~Che, Y.~Cheng, S.~Zhai, Z.~Sun, and Y.~Liu, ``Boosting deep learning risk
  prediction with generative adversarial networks for electronic health
  records,'' in \emph{2017 {IEEE} International Conference on Data Mining,
  {ICDM} 2017, New Orleans, LA, USA, November 18-21, 2017}, 2017, pp. 787--792.

\bibitem{kherif2003group}
F.~Kherif, J.-B. Poline, S.~M{\'e}riaux, H.~Benali, G.~Flandin, and M.~Brett,
  ``Group analysis in functional neuroimaging: selecting subjects using
  similarity measures,'' \emph{NeuroImage}, vol.~20, no.~4, pp. 2197--2208,
  2003.

\bibitem{hu2014convolutional}
B.~Hu, Z.~Lu, H.~Li, and Q.~Chen, ``Convolutional neural network architectures
  for matching natural language sentences,'' in \emph{Advances in Neural
  Information Processing Systems}, 2014, pp. 2042--2050.

\bibitem{lu2013deep}
Z.~Lu and H.~Li, ``A deep architecture for matching short texts,'' in
  \emph{Advances in Neural Information Processing Systems}, 2013, pp.
  1367--1375.

\bibitem{bordes2014open}
A.~Bordes, J.~Weston, and N.~Usunier, ``Open question answering with weakly
  supervised embedding models,'' in \emph{Machine Learning and Knowledge
  Discovery in Databases}.\hskip 1em plus 0.5em minus 0.4em\relax Springer,
  2014, pp. 165--180.

\bibitem{duchi2011adaptive}
J.~Duchi, E.~Hazan, and Y.~Singer, ``Adaptive subgradient methods for online
  learning and stochastic optimization,'' \emph{The Journal of Machine Learning
  Research}, vol.~12, pp. 2121--2159, 2011.

\bibitem{song2010observational}
J.~W. Song and K.~C. Chung, ``Observational studies: cohort and case-control
  studies,'' \emph{Plastic and reconstructive surgery}, vol. 126, no.~6, p.
  2234, 2010.

\bibitem{browner1988designing}
W.~S. Browner, S.~B. Hulley, and S.~R. Cummings, \emph{Designing clinical
  research: an epidemiologic approach}.\hskip 1em plus 0.5em minus 0.4em\relax
  Lippincott Williams \& Wilkins, 1988.

\bibitem{basu2004active}
S.~Basu, A.~Banerjee, and R.~J. Mooney, ``Active semi-supervision for pairwise
  constrained clustering.'' in \emph{SDM}, vol.~4.\hskip 1em plus 0.5em minus
  0.4em\relax SIAM, 2004, pp. 333--344.

\bibitem{rand1971objective}
W.~M. Rand, ``Objective criteria for the evaluation of clustering methods,''
  \emph{Journal of the American Statistical association}, vol.~66, no. 336, pp.
  846--850, 1971.

\bibitem{manning2008introduction}
C.~D. Manning, P.~Raghavan, H.~Sch{\"u}tze \emph{et~al.}, \emph{Introduction to
  information retrieval}.\hskip 1em plus 0.5em minus 0.4em\relax Cambridge
  university press Cambridge, 2008, vol.~1, no.~1.

\bibitem{meilua2007comparing}
M.~Meil{\u{a}}, ``Comparing clusterings—an information based distance,''
  \emph{Journal of multivariate analysis}, vol.~98, no.~5, pp. 873--895, 2007.

\bibitem{zhao2001criterion}
Y.~Zhao and G.~Karypis, ``Criterion functions for document clustering:
  Experiments and analysis,'' Citeseer, Tech. Rep., 2001.

\bibitem{van1974foundation}
C.~J. Van~Rijsbergen, ``Foundation of evaluation,'' \emph{Journal of
  Documentation}, vol.~30, no.~4, pp. 365--373, 1974.

\end{thebibliography}

\end{document}